%% file: main.tex
\documentclass{article}

\usepackage{PRIMEarxiv}

\usepackage[utf8]{inputenc} 
\usepackage[T1]{fontenc} 
\usepackage{hyperref} 
\usepackage{url} 
\usepackage{booktabs} 
\usepackage{amsfonts} 
\usepackage{nicefrac} 
\usepackage{microtype} 
\usepackage{fancyhdr} 
\usepackage{graphicx} 
\usepackage[numbers]{natbib} 
\usepackage{amsmath} 
\usepackage{amssymb} 
\usepackage{mathrsfs} 
\usepackage{comment} 
\usepackage{subcaption} 
\usepackage{graphicx} 
\usepackage{tikz} 
\usetikzlibrary{calc} 
\usetikzlibrary{arrows} 
\usepackage{pgfplotstable} 
\usepackage{pgfplots} 
\pgfplotsset{compat=1.18} 

\title{Adversarial Robustness for Deep Learning-based Wildfire Prediction Models
}

\author{
  Ryo Ide \\
  Visiting Scientist \\
  Department of Computer Science and Engineering \\
  University of Nevada, Reno, USA\\
  \texttt{ideryoryo@gmail.com}\\
   \And
  Lei Yang \\
  Associate Professor \\
  Department of Computer Science and Engineering \\
  University of Nevada, Reno, USA\\
  \texttt{leiy@unr.edu}\\
}

\begin{document}
\maketitle

\begin{abstract}
Rapidly growing wildfires have recently devastated societal assets, exposing a critical need for early warning systems to expedite relief efforts. Smoke detection using camera-based Deep Neural Networks (DNNs) offers a promising solution for wildfire prediction. However, the rarity of smoke across time and space limits training data, raising model overfitting and bias concerns. Current DNNs, primarily Convolutional Neural Networks (CNNs) and transformers, complicate robustness evaluation due to architectural differences. To address these challenges, we introduce WARP (Wildfire Adversarial Robustness Procedure), the first model-agnostic framework for evaluating wildfire detection models’ adversarial robustness. WARP addresses inherent limitations in data diversity by generating adversarial examples through image-global and -local perturbations. Global and local attacks superimpose Gaussian noise and PNG patches onto image inputs, respectively; this suits both CNNs and transformers while generating realistic adversarial scenarios. Using WARP, we assessed real-time CNNs and Transformers, uncovering key vulnerabilities. At times, transformers exhibited over 70\% precision degradation under global attacks, while both models generally struggled to differentiate cloud-like PNG patches from real smoke during local attacks. To enhance model robustness, we proposed four wildfire-oriented data augmentation techniques based on WARP’s methodology and results, which diversify smoke image data and improve model precision and robustness. These advancements represent a substantial step toward developing a reliable early wildfire warning system, which may be our first safeguard against wildfire destruction.
\end{abstract}


\section{Introduction}
Wildfires were among the deadliest US natural disasters in 2023, surpassing storms and floods \citep{Salas_2024}. {They are also some of the costliest, with the US} five-year average cost of firefighting estimated at USD 2.3 billion. Indeed, recent wildfires such as the 2025 Palisades Fires have demonstrated that uncontained fires may have a substantial cost on human life and infrastructure even in a short time frame~\citep{palisadesfire2025}. As wildfires become more frequent, early wildfire detection systems are imperative for evacuation and prevention.

Several works have proposed automated solutions with a particular focus on detecting early indications of wildfires. With rapid advancements in computer vision, image-based solutions have attracted substantial attention recently. Yazdi et al.~\cite{yazdi2022nemo} conducted a comparative study of existing wildfire monitoring approaches, including high- \citep{fernandes2004development} and low-altitude remote sensing \citep{govil2020preliminary}, local sensing \cite{barmpoutis2020review}, and terrestrial surveillance \citep{yazdi2022nemo}. They concluded that using deep learning with terrestrial surveillance to detect wildfire smoke was the most effective method of predicting wildfires. Terrestrial surveillance cameras positioned at vantage points have continuous temporal coverage with a wide field of view, producing high-resolution images. This allows wildfires to be detected in their incipient stage, the earliest stage of wildfires which is characterized by small smoke plumes. This makes terrestrial surveillance the most competitive monitoring approach.

\begin{figure}[bht!]
    \centering
    \input{figures/example_figure}
    \caption{Smoke object detection from surveillance video images. A DNN object detection model creates bounding boxes (see green box) to locate smoke as the target object. Image adapted from \cite{ALERTWildfire}.}
    \label{fig: OD_illustration}
\end{figure}
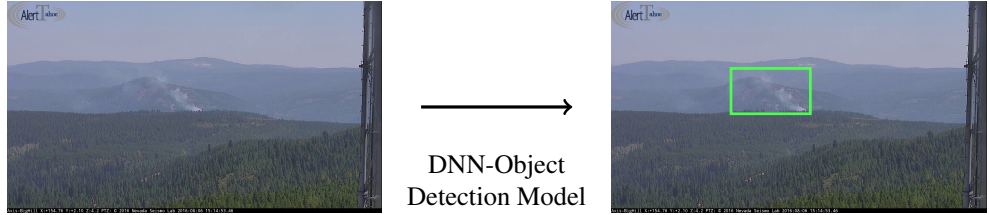

Wildfire smoke detection using deep learning typically involves object detection, a highly established task in computer vision. This consists of identifying and locating an object with a bounding box within an image. The target object is wildfire smoke in this context. Figure~\ref{fig: OD_illustration} illustrates wildfire smoke object detection from surveillance video images using deep learning. To develop object detection models, large-scale deep neural network (DNN)
models are trained on image data, usually in quantities well over the thousands for each target object. Using the trained DNN models, smoke objects can be detected in real time, creating a powerful synergy with continuous surveillance. By doing so, automatic prediction systems can be developed which can (1) replace human resources that are expensive and time-consuming, and (2) potentially detect with a higher precision than their human counterparts. In tasks like wildfire detection, which is both costly~\cite{NIFC}~{and demands accuracy (see Section }\ref{subsec: challenges}), these advantages of DNNs are particularly desirable.

\subsection{Existing DNN Solutions to Wildfire Smoke Detection}
\label{subsec: existing solutions}

Existing camera-based wildfire smoke detection using computer vision frameworks in the literature are characterized by two major DNN architectures: \textit{CNN} and \textit{transformer}. There are numerous CNN-based approaches~\citep{goncalves2024, jindal2021real,jeong2020light, fernandes2022,al2023early};~in particular,  Jeong et al.~\cite{jeong2020light} proposed a real-time wildfire detection model, where a YOLOv3~\citep{redmon2018yolov3}-based CNN framework (for producing candidate proposals for smoke-positive regions) is combined with Long-Short Term Memory (LSTM) \citep{hochreiter1997long} (for screening the candidates).  While enforcing temporal consistency is demonstrated to be effective in distinguishing smoke and cloud, the model was trained on samples taken from only 12 scenes. Despite a large number of image frames, the effective sample size may be as few as 12, potentially limiting the model's ability to generalize.
Additionally, for wildfire \textit{flame} detection, which is considerably easier than \textit{smoke} detection, Al-Smadi et al.~\cite{al2023early} developed another CNN-based detection model using newer versions of YOLO to report an extremely high prediction performance. However, the model was trained on positive-only samples with a relatively small image count (1723). It is not entirely clear how the limited sample size and synthetically inflated wildfire occurrences affect model robustness and generalizability.

{For the transformer approach, there have only been two camera-based transformer approaches to wildfire detection in the literature, to the best of our knowledge}~\cite{yazdi2022nemo,wang2023efficient}. ~{In particular,} Yazdi~et al.~\cite{yazdi2022nemo} proposed NEMO (NEvada sMOke detection benchmark), using the DEtection TRansformer (DETR) \citep{carion2020end} framework for the first time in wildfire smoke detection. They set the practical benchmark for state-of-the-art precision.
 To address the issue of false positives, they use several data preprocessing strategies to artificially diversify data. Specifically, they inject 260 smoke-negative background images randomly selected from the internet. They also create 116 collages 
from reused smoke-positive and smoke-negative images to further address this problem. Generally, the collages were effective not only for reducing false positives but for introducing variety to the object's position and size. It was shown that this preprocessing step can reduce the false alarm rate. However, detailed discussions on the model's generalizability and robustness on real-life data are not present beyond an ablation study of the model's detection encoder mechanism and a typical time-series detection analysis.

\subsection{Challenges in Adapting DNN Solutions to Wildfire Smoke Detection}
\label{subsec: challenges}

While the DNN-based object detection approach has shown promising results~\citep{al2023early,jeong2020light,yazdi2022nemo}, two main challenges remain: (1) the limited sample size coupled with low 
diversity in wildfire smoke training images, and (2) the lack of universal solutions for improving model robustness.

The first challenge arises because of the properties of wildfire smoke. Smoke is \textit{spatially} anomalous because it occupies a small portion relative to the entire image. As a result, it is absent from the vast majority of pixels in the frames of high-resolution continuous surveillance videos. This makes manual annotation of images for object detection highly labor-intensive, necessitating ad hoc preprocessing steps as discussed in Section~\ref{subsec: existing solutions}. Furthermore, smoke is \textit{temporally} anomalous because it originates from a rare event. This results in surveillance image sequences having relatively small smoke-positive frames compared to the entire data. {Thus, limited data generally create imbalance problems,}~\citep{cho2020transfer}~{and when used for training, can cause detrimental performance degradation}~\cite{buda2018systematic,johnson2019survey}.~{Furthermore, w}hen trained on limited datasets, DNNs are known to produce unexpected and/or detrimental outcomes even from slight modifications to the input \citep{goodfellow2014explaining}.  In wildfire smoke detection, where training data are almost always limited, this problem can be exacerbated by the current trend of shifting from CNNs (Convolutional Neural Networks) to transformer-based models~\citep{han2022survey},~{as the latter explicitly handles second-order statistics through the key-value transformation }~\citep{dosovitskiy2020image}.
Despite the importance of this issue, little work has been done to assess the robustness of wildfire detection models across various inputs.   

{The second challenge arises because DNNs are inherently \textit{black boxes}}~\cite{Molnar_2024a}. Since the internal workings of the model, including feature extraction processes and the role of parameters, are not immediately discernible, addressing potential issues requires substantial case-by-case effort, even if the source code is available. The architecture of CNNs and transformers differs greatly, making analysis based on model-specific quantities (e.g.,~gradients) less practical. Since the spatiotemporal anomaly issue could potentially introduce vulnerabilities to the model, a model-agnostic framework for evaluating the model's robustness and fine-tuning must be developed. 

\subsection{Contributions}

We recognize the following advantages of DNNs for wildfire detection: previous studies have found that DNN--computer vision models are the most effective in wildfire detection due to their compatibility with surveillance cameras, which have the most coverage compared to other wildfire observation methods. They also synergize well with cameras because they can keep up with their footage in real time, allowing for an automated wildfire detection system that saves costs and human errors.

However, we also recognize the following disadvantages of DNNs for wildfire detection: because of smoke's fundamental properties, smoke image data is highly difficult to collect, which not only creates a severe data shortage but also a class imbalance problem. This limits the model's ability to be generalized to all wildfire scenarios. More importantly, DNNs are black boxes, meaning their internal decision-making mechanisms are too abstract to interpret. This makes it challenging to identify specific sources of robustness vulnerabilities. Not only that, these mechanisms differ greatly across architectures, making comparing robustness across different models difficult. 

To address these challenges, we propose WARP (Wildfire Adversarial Robustness Procedure), the first model-agnostic framework to comprehensively evaluate the adversarial robustness of wildfire detection models. Unlike common adversarial attack methods, which often rely on solving an optimization problem for perturbations using internal model details, WARP is model-agnostic and uses relatively simple noise injection methods. Additionally, WARP incorporates wildfire-specific contexts in designing noise tests, especially in the distinction between smoke and cloud, rather than using generic random noise. To the best of our knowledge, WARP is the first framework to offer a model-agnostic, contextual adversarial robustness evaluation method specifically tailored for wildfire detection. Insights obtained through WARP's analysis can be used to further improve the model through data augmentation, making it a crucial first step toward a truly practical/reliable wildfire detection system.
The main contributions of our study are summarized as follows:
\begin{itemize}
  \item We propose WARP, consisting of global and local model-agnostic evaluation methods for model robustness, tailored to wildfire smoke detection. 
  \item We compare the robustness of DNN-based wildfire detection models across two major neural network architectures, namely CNNs and transformers, and provide detailed insights into specific vulnerabilities of those models. 
  \item We propose data augmentation approaches for potential model improvement based on the above findings.
\end{itemize}


\section{Preliminaries}

We seek to quantitatively evaluate the adversarial robustness of wildfire smoke detection models, comparing the two main DNN architectures: CNN and transformer. This section formally summarizes the problem setting and provides a concise overview of adversarial robustness.

\subsection{Problem Statement}

We assume that an image-based wildfire detection model $y=f(x)$ is given, where $x$ is an input image, typically an image frame captured by a surveillance camera, and $y \in \{0,1\}$ is the binary label indicating whether $x$ contains wildfire smoke ($y=1$) or not ($y=0$). Since wildfire smoke is spatially localized, the function $f$ incorporates two steps: bounding box generation (locating subimages within $x$) and scoring (estimating the probability of containing smoke), as illustrated in Figure~\ref{fig: OD_illustration}.

Additionally, we assume that a test dataset $\mathcal{D}= \{x^{(1)}, \ldots, x^{(N)}\}$ is available, where $x^{(n)}$ is the $n$-th image. Since the input images are often noisy due to varying weather and lighting conditions, a practical wildfire detection system must be robust to small variations in test images. The central question we address in this study is \textit{how robust} such a wildfire detection model is.  

\begin{figure}[hbt!]
    \centering
    \colorbox{gray!5}{ 
        \includegraphics[width=0.7\linewidth]{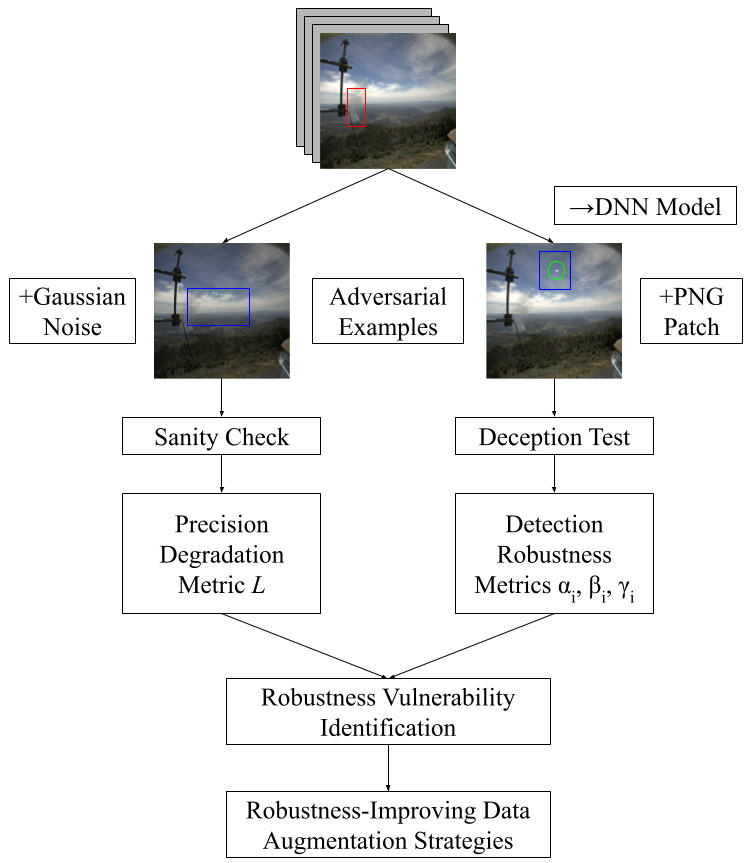}
    }
    \caption{WARP workflow.}
    \label{fig: warp workflow}
\end{figure}

We address this question by proposing metrics indicating \textit{changes} in the classification outcomes upon introducing small perturbations to the input image. Specifically, as illustrated in Figure~\ref{fig: warp workflow}, we compute 
the precision degradation metric $L$, the classification flip probabilities $\alpha_i$ and $\beta_i$, and the localization deception rate $\gamma_i$. $L$ (see Section \ref{subsec: global sanity check: noise overlay}) seeks to quantify the model's robustness against global noise, whereas $\alpha_i$, $\beta_i$, and $\gamma_i$ (see Section \ref{subsec: local deception test: noise patch} seek to quantify the model's robustness against local noise. 

As discussed, wildfire smoke detection models need an adaptable and contextualized adversarial robustness evaluation method. For a disaster prevention task where human lives are involved, wildfire detection solutions must be \textit{robust}. That is, a DNN wildfire prediction model must be able to adapt to a variety of wildfire scenarios, not exclusive to the wildfires it has seen in its training data.  We leverage adversarial attacks to identify any critical model vulnerabilities through a variety of noise perturbations. The insights obtained will be vital, especially in a task like image-based wildfire smoke detection, as they can be used to further diversify the training data through data augmentation which will be discussed in Section \ref{sec: Discussion}.

\subsection{Adversarial Robustness}
\textit{Adversarial robustness} refers to a model's resilience to external attacks on its input. ``Attacks'' usually involve the injection of \textit{noise perturbations} (slight modifications in the form of noise) into the original input to artificially create counterfactual (unforeseen) test scenarios. These are called \textit{adversarial examples}.

In a classification task, adversarial robustness is typically evaluated by quantifying the resilience of the original detection when facing input perturbations. Let $f(x)$ be the classification model, $f:\mathbb{R}^{m}\longrightarrow \left\{ 1,...,k \right\}$, where $m$ is the dimension of the input image tensor ($\text{width}\times\text{height}\times\text{color channels}$), $k$ is the number of classes. In wildfire smoke detection, $k$ is 2 as the only classes are \textit{null} and \textit{smoke}. The fundamental question that is considered when evaluating adversarial robustness is as follows:

\begin{quote}
Given a test input $x \in \mathbb{R}^m$ and a perturbation $r \in \mathbb{R}^m$, compare how $f(x)$ differs from $f(x+r)$. 
\end{quote}

If $f(x)$ matches the ground truth, but $f(x+r)$ does not, the model cannot be considered robust. However, it is conditionally required that $r$ be reasonably small, as it is almost guaranteed that the classification output will change if $r$ is arbitrarily large. Therefore, the most desirable perturbed image tensor $x' \triangleq x +r$ looks almost identical to the original image tensor $x$, but has the maximum impact on classification outcome. The optimization problem to find such a perturbation is called an \textit{adversarial attack} \citep{Molnar_2024a}:
\begin{align}\label{eq:Adversarial_Robustness_general}
    \max_{r} \mathrm{Loss}(x+r, l) \quad \mbox{subject to}\quad \mathrm{Distance}(x,x+r) \leq \epsilon
\end{align}
where $l$ is the predicted class label for $x$ (i.e., $f(x) = l$), Loss$(x,y)$ is a loss function for an input and output pair $(x,y)$, and $\epsilon$ is a small positive error constant to keep the perturbed input close to the original input. Typically, the loss function is the same as that of the trained model. 

Depending on their nature, perturbations can generally be categorized into two categories: global and local attacks \citep{chen2022adversarial}. \textit{(1) Global Attacks}: The authors of \cite{szegedy2013intriguing,goodfellow2014explaining} proposed global attack methods that seek the optimal perturbation that covers the entire input image. The authors of \cite{szegedy2013intriguing} incorporated the distant constraint as a penalty term $c\|r\|_1$, where $c$ is a constant and $\|\cdot\|_1$ is the $\ell_1$ norm, that is added to the objective function. Moreover, \cite{goodfellow2014explaining} proposed a more computationally efficient approach by constraining $r$ not by a constant but only to the sign of the gradient of the loss function $\frac{\partial}{\partial x}\mathrm{Loss}(x,y)$.
\textit{(2) Local Attacks}:
The authors of \cite{brown2017adversarial,su2019one} proposed local attack methods. \cite{su2019one} proposed a single-pixel attack method based on a differential evolution framework. However, single-pixel attacks are highly inefficient for the typical object detection input which is a $640\times640$ pixel image. Additionally, \cite{brown2017adversarial} proposed a patch-based method, where a constant patch would be directly injected into the image. These patches adapt to different backgrounds and transformations, including position, size, and rotation. Patches were generally successful at deceiving otherwise rigorously trained models, thus making them a prime tool for adversarial robustness.

There are several trends for both categories of existing adversarial attack approaches that limit their utility when applied to wildfire smoke detection. 

Global attack methods often require model-specific elements for efficient computation, such as the gradient of the loss function. This necessitates the full knowledge of the architecture of the DNN model, thus making each model's adversarial attack case-by-case. Given the variety of DNN architectures in wildfire smoke detection and the rapid development of new computer vision frameworks, this reliance on model elements may be computationally expensive. To comprehensively evaluate adversarial robustness for all wildfire smoke detection frameworks, a model-agnostic approach is necessary. 

Moreover, local attack methods typically disregard the \textit{context} of their task. Both single-pixel attacks and patch noise attacks are generally abstract formulations when viewed by humans, and generally do not fit in the context of wildfire smoke detection (i.e., abstract glob-like patches will not appear in camera surveillance images). Thus, their utility may be limited to deceiving {hyper-tuned models} in highly specific conditions. {Wildfire smoke detection demands a more nuanced usage for noise patches. C}ontextual cues, such as the subtle differences in objects' shape, color, and position, are paramount to distinguish smoke from {highly similar objects (e.g., clouds, man-made structures with similar coloration) at a considerable distance. With limited wildfire data, it is not an option to make models learn these contextual cues by retraining them on additional images.} Therefore, patch noise attack methods must take context into account to (1) evaluate a model's ability to differentiate objects using the task's context and (2) create adversarial examples that train this ability.


\section{Wildfire Adversarial Robustness Procedure}

In this section, we introduce the proposed framework, WARP (Wildfire Adversarial Robustness Procedure), for comprehensively assessing the robustness of smoke detection models.  WARP generates two types of adversarial examples through data augmentation: global noise (i.e., Gaussian noise) and local noise (i.e., cloud PNG patches).  We then compare the original detection (red bounding box) with the perturbed detection (blue bounding box): we conduct a sanity check for the images' perturbed with global noise, which analyzes the effect of global noise on model precision (see Section \ref{subsec: global sanity check: noise overlay}). Additionally, we conduct the deception test for images perturbed with local noise, which analyzes the robustness of the model's localization and classification abilities on specific objects (see Section \ref{subsec: local deception test: noise patch}). We later demonstrate in Section \ref{sec: Discussion} that the tests outlined above can identify vulnerabilities in robustness for wildfire detection models, which can produce data augmentation strategies tailored specifically for improving the robustness of DNN models. 

\subsection{Global Sanity Check: Noise Overlay}
\label{subsec: global sanity check: noise overlay}

For practical wildfire smoke detection, models must distinguish smoke from similar objects such as clouds, fog, and camera artifacts. Since this distinction can be subtle, model training typically requires a significant sample size, which is currently unavailable in wildfire smoke detection, as discussed in Section~\ref{subsec: existing solutions}. This raises questions about whether existing DNN models are robust enough against adversarial attacks.

As a preliminary test, the \textit{global sanity check} uses image-wide random perturbations. Specifically, a random noise overlay following the Gaussian distribution ($r \sim \mathcal{N}(0, 1)$, i.e., Gaussian noise with zero mean and unit variance) of the same size as the input tensor $x$ is added to the entire image. The perturbed image $x'$ is given by
\begin{equation}
x' = (1-a)x + a\sigma r,
\label{eq: noise overlay def}
\end{equation}
where $a$ denotes the noise level that takes a value in the range $[0,1]$, and $\sigma$ is the standard deviation of the input image tensor. The perturbation is contextualized to the input image since $\sigma$ is uniquely computed from that particular image.

For a given noise level, the \textit{mAP (mAP50-95) percentage loss} $\mathcal{L}$ is calculated to quantify the precision loss from the addition of random noise. We choose \textit{mAP} (see Appendix \ref{subsec: map} for detailed definitions) as the primary challenge metric since fixed threshold-precision metrics (i.e., mAP33, mAP50, mAP75, etc.) may only offer a limited evaluation of precision. \textit{mAP} measures the mean average precision across Intersection over Union (IoU) thresholds ranging from 0.50 to 0.95, making it a robust metric \citep{yazdi2022nemo}.
\begin{equation}
\mathcal{L}=\frac{mAP_\text{after}-mAP_\text{original}}{mAP_\text{original}} \times 100,
\label{eq: map percentage loss}
\end{equation}
where $mAP_{\text{after}}$ is the mAP score after adding random noise to the dataset, and $mAP_\text{original}$ is the mAP score for the original dataset.

\subsection{Local Deception Test: Noise Patch}
\label{subsec: local deception test: noise patch}

\begin{figure}[bth]
    \centering
    \begin{minipage}{1\textwidth}
    \captionsetup{width=1\textwidth} 
    \centering
    \begin{subfigure}[b]{0.3\textwidth}
        \captionsetup{width=\textwidth} 
        \centering
        \includegraphics[width=0.7\textwidth]{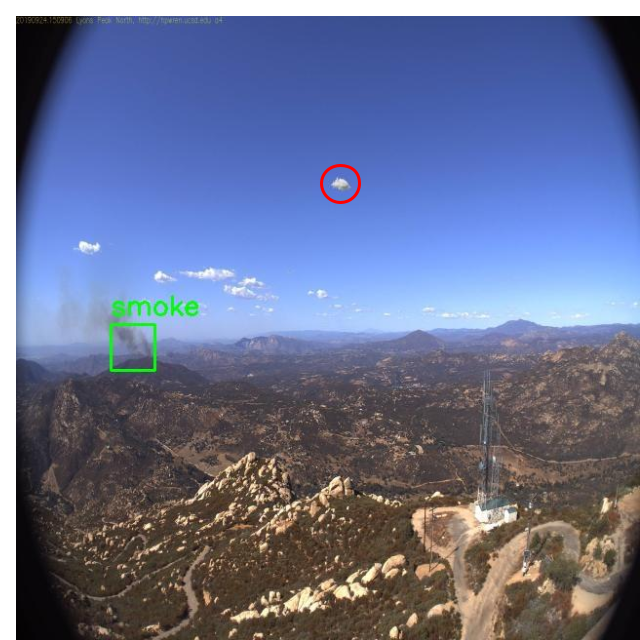}
        \caption{}
        \label{fig: example augmented image}
    \end{subfigure}
    \begin{subfigure}[b]{0.3\textwidth}
        \captionsetup{width=\textwidth} 
        \centering
        \includegraphics[width=0.4\textwidth]{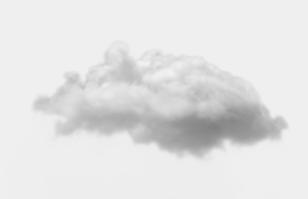} 
        \caption{}
        \label{fig: cloud png patch}
    \end{subfigure}
    \caption{Local noise injection. (a) An example image with injected cloud-like noise at a grid location, highlighted by the red circle. The green bounding box indicates the ground-truth location of the smoke. (b)~The cloud-like PNG patch used as local noise, with a background added for visibility. Adapted from the internet.}
    \label{fig: noise patch details}
    \end{minipage}
\end{figure}

The previous test considers only the context of noise level variability via $\sigma$. We also propose the \textit{local deception test}, which introduces \textit{spatial} context specific to the smoke detection task. Specifically, local noise patches (small images in the PNG (portable network graphics) format) are used to check for robustness against localized perturbations specific to wildfires. For each image $i$, a PNG patch of constant size is injected at a specific spot in the image (see Figure \ref{fig: noise patch details}a for instance). For computational efficiency, we divide each image into 25 by 25 grids, and the noise is injected in the center of each grid slot. ``Noise” can be any wildfire-related object, including smoke-like objects such as clouds, or other objects in context, such as trees, buildings, glare, etc. We specifically used clouds as they are the most common subject of false positives in previous works \citep{jeong2020light, al2023early, yazdi2022nemo}. Figure~\ref{fig: noise patch details}b shows the cloud PNG used in the local deception test. Despite being horizontally wider than wildfire smoke, they represent the everyday {cumulus}-type clouds, which is why we selected them. After observing existing wildfire smoke data, and testing different configurations, we configured the patch to be $25\times 25$ pixels at 100\% brightness.

To quantify the robustness against patch noise, three metrics are proposed, $\alpha_i, \beta_i$, and $\gamma_i$. They are defined in Equations (\ref{eq:alpha})--(\ref{eq: deception rate}) in the following sections. 

\subsubsection{Classification Flip Probabilities}

A wildfire smoke detection model's ability can be subdivided into its classification 
 and localization 
 abilities. The classification ability refers to how precisely the model can identify an image as smoke-positive or smoke-negative. The localization ability refers to how precisely the model can locate the smoke object using bounding boxes given a smoke-positive image. 

The Classification Flip Probabilities 
 focuses on the vulnerability of the model's classification abilities. When smoke-positive input images are given, a model may fail to generate bounding boxes for some images. These misclassified images can be called false negatives (FNs). On the other hand, if the model successfully generates at least one bounding box in any location of an image, the image can be called a true positive (TPs).

The Classification Flip Probabilities quantify how many TPs or FNs are ``flipped'' upon injecting \textit{local} noise. Unlike the global sanity check test, noise injection is performed grid-wise. Image $i$ is divided into $A_i=25\times 25=625$ equal grids, and local noise is injected into each grid slot. The classification outcome is then observed $A_i$ times in total for the image. For each image $i$, depending on whether it is a TP or FN, the classification vulnerability of a model can be quantified by calculating two conditional probabilities: 
\allowdisplaybreaks
\begin{align} 
\alpha_i &= P_i(l_\text{null} \mid l_\text{smoke})
= \frac{1}{A_i}\sum_{j=1}^{A_i} \mathbb{I}_{i,j}(\text{FN} \leftarrow \text{TP}),\label{eq:alpha}
\\
\ \beta_i &= P_i(l_\text{smoke} \mid l_\text{null})
= \frac{1}{A_i}\sum_{j=1}^{A_i} \mathbb{I}_{i,j}(\text{TP} \leftarrow \text{FN}), \label{eq:beta}
\end{align}
where $l_\text{null}$ and $l_\text{smoke}$ denote the null and smoke classes. $\mathbb{I}_{i,j}(\cdot)$ is the indicator function, which equals 1 if the specified flip occurs for image $i$ when the noise is injected into the $j$-th grid slot. It equals 0 otherwise. Again, $A_i$ is fixed at $25\times 25 = 625$ possible slots for every~image.

These image-wise probabilities are averaged over the entire test set data to evaluate the model's classification vulnerabilities. To summarize what these metrics mean, $\alpha_i$ and $\beta_i$ represent the classification robustness that image $i$ is smoke-positive and smoke-negative,~respectively.

\subsubsection{Localization Deception Rate}

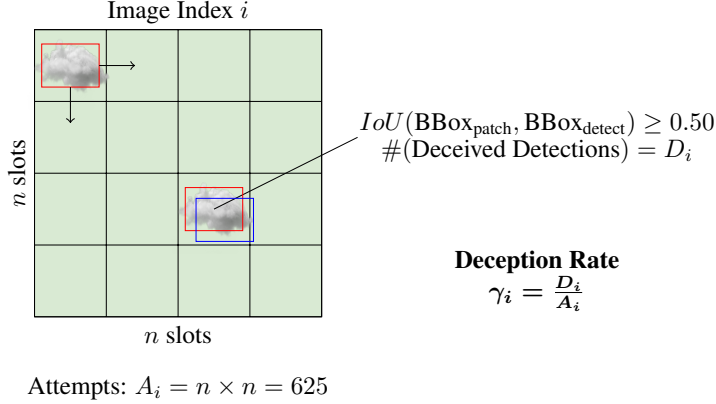
\begin{figure}[hbt!]
    \centering
    \resizebox{0.6\textwidth}{!}{\input{figures/noise_workflow}}
    \caption{Illustration of the proposed localization deception test.}
    \label{fig: deception test workflow}
\end{figure}

 We propose the \textit{Localization Deception Rate} $\gamma_i$ (for each image $i$ in the test set data) to evaluate the localization ability. Unlike the Classification Flip Probabilities, $\gamma_i$ quantifies bounding box detection robustness, which is defined below (see Figure \ref{fig: deception test workflow}): 
\begin{equation}
\gamma_{i}=\frac{D_{i}}{A_{i}},
\label{eq: deception rate}
\end{equation}
where $D_{i}$ is the number of detections and its bounding box has $IoU\geq0.50$ (see Equation~\eqref{eq: IoU} for further details) with the injected noise's bounding box. $A_{i}$ is the number of attempts, or the number of possible positions for the noise to be injected in a $25\times25$ grid ($25^{2}=625)$. Since $D_{i}$ is discrete, $\gamma_{i}$ is also discrete. That is, $\{1, 2, 3,\dots, 625\}$ is the set of all possible values of $D_{i}$, and the set of all possible deception rates $\gamma_{i}$ is $\{1/625, 2/625, 3/625,\dots,625/625\}=\{0.0016, 0.0032, 0.0048,\dots, 1.0000\}$.

A higher $\gamma_{i}$ indicates that the model is more vulnerable to noise injection for image $i$. The overall localization robustness of the model can be evaluated by averaging $\gamma_{i}$ across all images in the test set data. In a theoretical model with perfect adversarial robustness, all values of $\gamma_{i}$ should have a frequency of 0.

\subsection{Data Collection and Preparation}

We used two datasets to train, validate, and test the model: the NEMO dataset \citep{yazdi2022nemo}, curated from various sources including \citep{ALERTWildfire}, and a dataset from \cite{govil2020preliminary}, collected from the High-Performance Wireless Research and Education Network (HPWREN) database  \citep{HPWREN}.

The NEMO dataset was originally created to fine-tune DETR. It contains 2500+ images, with roughly 90\% being smoke-positive. The smoke-positive regions in the original video frames were cropped and zoomed for efficient training. Following \cite{yazdi2022nemo}, we used this dataset for model training. In contrast, the dataset by \cite{govil2020preliminary} contains more realistic 1661 unedited smoke-positive images. 
We used this dataset for model testing and robustness evaluation. Table~\ref{tab: data split} shows the data split.

\begin{table}[hbt!]
    \centering
    \caption{Data split for the image dataset. Training and validation data were sourced from the NEMO \protect\citep{yazdi2022nemo}, whereas the testing data were sourced from HPWREN \protect\citep{govil2020preliminary}.}
    \label{tab: data split}
    \begin{tabular}{c c c }
        \hline\hline
        \textbf{Training} & \textbf{Validation} & \textbf{Testing}\\
        \hline
         2,704 &  337 & 1,661\\
        \hline
    \end{tabular}
\end{table} 

\subsection{Object Detection Framework}
\label{subsec: object detection framework}

Since the ultimate goal is to establish an automated wildfire smoke detection system allowing real-time detection and continuous model improvement, we focus on object detection models allowing \textit{real-time} object detection. Real-time models allow faster feedback when detecting real-life adversarial scenarios. Additionally, we contributed to the nascent model zoo in camera-based real-time wildfire smoke detection \citep{jeong2020light, al2023early, yazdi2022nemo, wangmachinevision, jindal2021real, oh2020early, wei2022intelligent} by introducing two previously unused open-source models to the field. 

\begin{itemize}
    \item For a CNN-based real-time object detection model, we choose YOLOv8, the 8th generation model of the YOLO (You Only Look Once) framework. It is a popular real-time object detection framework and is publically available on the Ultralytics API \citep{Ultralytics_2024}.

    \item For a transformer-based real-time object detector, we choose RT-DETR (Real-Time Detection-Transformer), which can be viewed as a real-time variant of DETR used by \cite{yazdi2022nemo}. RT-DETR overcomes the computation-costly limitations of transformers by sacrificing minimal accuracy for speed by prioritizing and selectively extracting object queries that overlap the ground truth bounding boxes by a certain IoU \citep{zhao2024detrs}.
\end{itemize}

Pre-trained weights were transferred from COCO-dataset-trained versions of YOLOv8 and RT-DETR (both v8.3.67). Their lightweight versions, YOLOv8-nano (YOLOv8n) and RT-DETR-large (RT-DETR-l), were chosen to reduce computation time during robustness evaluation. Default training and inference hyperparameters were used from \citep{Ultralytics_2024}. {Training YOLOv8n for 250 epochs and RT-DETR-l for 285 produced our best results. However, e}xtensive hyperparameter tuning is encouraged in the future. 

We use metrics mAP and mAP50 to measure model precision (See~Appendix \ref{subsec: map} for detailed definitions), which are common challenge metrics in object detection. All training, validation, and testing were run on a Python~3.8.12 virtual environment with CUDA version 12.2, equipped with quadruple NVIDIA GeForce GTX 1080 GPUs from the University of Nevada, Reno.

\begin{table}[bht!]
    \centering
    \caption{Training arguments. Other hyperparameters follows default values from \citep{Ultralytics_2024}.}
    \label{tab: train args}
    \begin{tabular}{c c c}
    \hline\hline 
        \textbf{Hyperparameter} & \textbf{YOLOv8n} & \textbf{RT-DETR-l}\\
        \hline
        epoch & 250 & 285\\
        \hline
        batch & 32 & 32\\
        \hline
        initial learning rate
        & 0.01 & 0.01\\
        \hline
    \end{tabular}
\end{table}


\section{Results}

This section presents the results of the adversarial robustness of CNN- and transformer-based wildfire detection model architectures, trained on two publicly available datasets.

\subsection{Post-Training}
Table \ref{tab: training results} compares precision metrics for \textit{real-time} YOLOv8n and RT-DETR-l. Metrics for \textit{non-real-time} models are also shown for comparison, which are the transformer-based model (NEMO-DETR) and last-generation CNN-based models (NEMO-FRCNN, NEMO-RNet), trained by \cite{yazdi2022nemo}. YOLOv8n outperforms RT-DETR-l in precision and parameter efficiency. Specifically, YOLOv8n achieved a very efficient mAP-to-parameter ratio with little hyperparameter tuning. YOLOv8n approached NEMO-DETR's mAP by $\approx$7.80\%, whereas RT-DETR-l approached it by $\approx$23.6\%. YOLOv8n and RT-DETR-l outperformed NEMO-FRCNN and NEMO-RNet, but lost to the state-of-the-art NEMO-DETR. We expected this as YOLOv8n and RT-DETR-l had a reduced parameter size for computational efficiency. Nevertheless, both are competitive wildfire smoke detection models.

\begin{table}[hbt!]
    \centering 
    \caption{Comparison of smoke detection accuracies. Note that NEMO models are not real-time and for comparison purposes only. Their results were adapted from~\citep{yazdi2022nemo}.}
    \label{tab: training results}
    \begin{tabular}{c c c c c}
    \hline \hline
    & \textbf{mAP} 
& \textbf{mAP50}
& \textbf{Parameter size}
& \textbf{mAP-to-param ratio}\\
    \hline
    \textbf{{YOLOv8n}} & {39.0} & {72.0} & {3.2M} & {$2.25\times10^{-5}$}\\
    \hline
    \textbf{{RT-DETR-l}} & {32.2} & {69.7} & {33M} & {$2.11\times10^{-6}$}
    \\ \hline 
    \hline 
    \textit{NEMO-DETR} & 42.3 & 79.0 & 41M & $1.93\times10^{-6}$\\
    \hline
    \textit{NEMO-FRCNN} & 29.5 & 69.3 & 43M & $1.61\times10^{-6}$\\
    \hline
    \textit{NEMO-RNet} & 28.9 & 68.8 & 32M & $2.15\times10^{-6}$\\
    \hline
    \end{tabular}
\end{table}

Despite its state-of-the-art transformer architecture, RT-DETR-l's precision was lower than the CNN-based YOLOv8n. This may be attributed to RT-DETR-l's IoU-Aware Query Selection mechanism \citep{zhao2024detrs}. This mechanism prioritizes object queries with good initial IoU with ground truth. Small objects like wildfire smoke have small bounding boxes, making it likely that their IoU with initial predictions is low. Object queries for small smoke are therefore less prioritized during training, making it harder for RT-DETR to pick up smoke's subtle features. Without extensive data fine-tuning and hyperparameter-tuning, speed-optimized transformers may be unable to take advantage of the strengths of the self-attention mechanism. This requirement can be harmful for nuanced tasks that require continual updating to accommodate unforeseen circumstances. Wildfire smoke detection may be only one such case where this may be a problem.

In addition, general model evaluation approaches using metrics that do not account for model robustness may not be appropriate. Modern metrics reflect the model's performance under one validation dataset. Even if extensive hyperparameter tuning can significantly improve precision, it may not noticeably change results in practical testing.

\subsection{Results of Global Sanity Check}
\label{subsec: results of global sanity check}

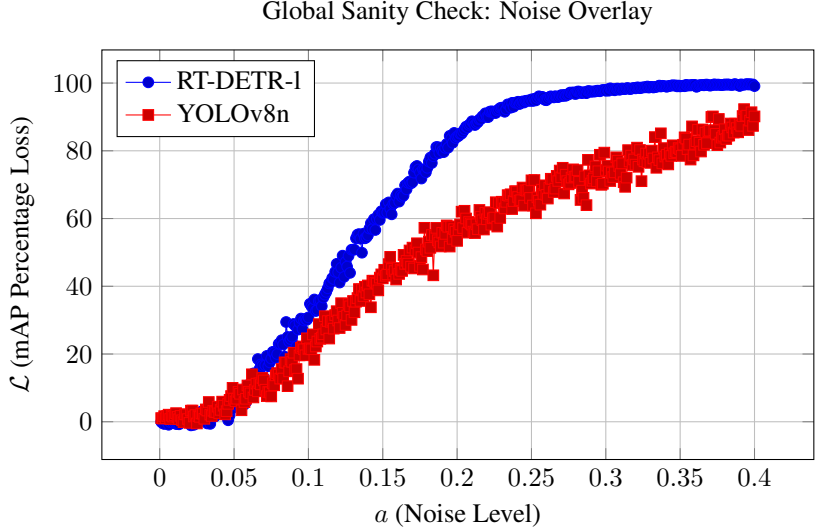
\begin{figure}[hbt!]
    \centering
    \begin{tikzpicture}
        \begin{axis}[
            width=11cm,
            height=7cm,
            xlabel={$a$ (Noise Level)}, 
            ylabel={$\mathcal{L}$ (mAP Percentage Loss)}, 
            title={Global Sanity Check: Noise Overlay},    
            grid=both, 
            legend style={at={(0.160, 0.975)},
            anchor=north,
            legend columns=1}, 
            legend cell align={left},   
            scaled x ticks=false,  
            tick label style={/pgf/number format/fixed}, 
            xtick={0, 0.05, 0.10, 0.15, 0.20, 0.25, 0.30, 0.35, 0.40}  
        ]
        \addplot table{data/rtdetr_map_loss_normalized.txt}; 
        \addlegendentry{RT-DETR-l}
        \addplot table{data/yolov8_map_loss_normalized.txt}; 
        \addlegendentry{YOLOv8n}
        \end{axis}
    \end{tikzpicture}
    \caption{mAP Percentage Loss plotted across all iterations.}
    \label{fig: noise overlay sanity check}
\end{figure}

Figure~\ref{fig: noise overlay sanity check} shows the result of the global sanity check as the noise level $a$ is increased in increments of 0.001 from 0.0 to 0.4 over 400 iterations. Interestingly, we observed a substantial difference between the two architectures. Specifically, the mAP percentage loss $\mathcal{L}$ for the transformer-based RT-DETR-l generally degraded at a higher rate than that of the CNN-based YOLOv8n. $\mathcal{L}$ for RT-DETR-l converged to $\approx$100 at noise level $a\approx0.300$, whereas YOLOv8n did not show signs of convergence even after the terminal noise level 0.400 ($\mathcal{L}_\text{RT-DETR-l}(0.400)=99.23$ and $\mathcal{L}_\text{YOLOv8n}(0.400)=89.53$).

\begin{figure}[hbt]
    \centering
    \begin{minipage}{0.47\textwidth}
    \captionsetup{width=1\textwidth} 
    \centering
    \begin{subfigure}[b]{0.48\textwidth}
        \centering
        \includegraphics[width=\textwidth]{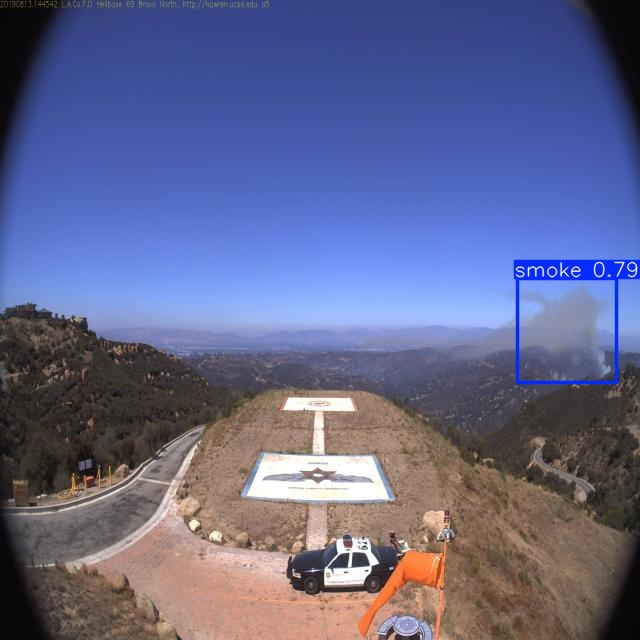}
        \caption{}
        \label{fig: yolo global scene1a}
    \end{subfigure}
    \begin{subfigure}[b]{0.48\textwidth}
        \centering
        \includegraphics[width=\textwidth]{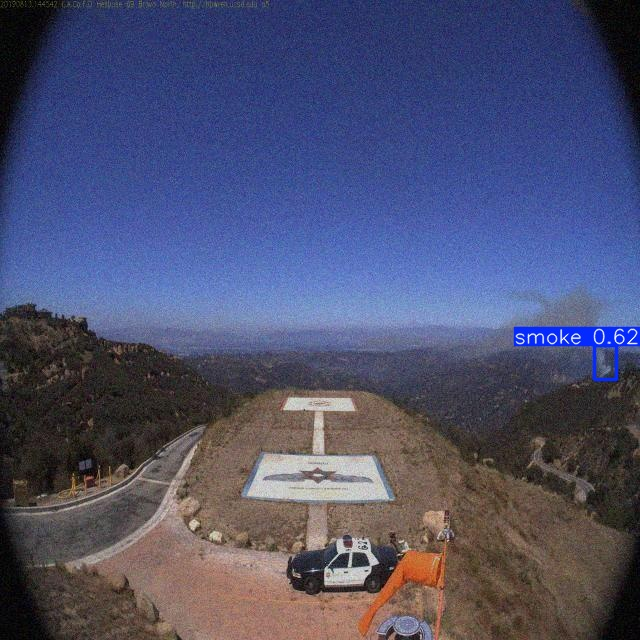} 
        \caption{}
        \label{fig: yolo global scene1b}
    \end{subfigure}
    \caption{(a) YOLOv8n without noise; detection confidence 0.79 and (b) with noise; detection confidence 0.62.}
    \label{fig: yolo global scene1}
    \end{minipage}
    \hspace{4mm}
    \begin{minipage}{0.47\textwidth}
    \captionsetup{width=1\textwidth} 
    \centering
    \begin{subfigure}[b]{0.48\textwidth}
        \centering
        \includegraphics[width=\textwidth]{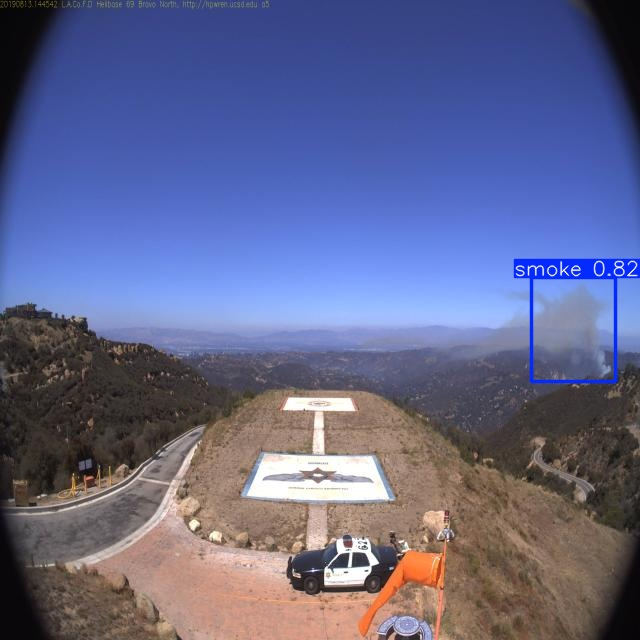} 
        \caption{}
        \label{fig: rtdetr global scene1a}
    \end{subfigure}
    \begin{subfigure}[b]{0.48\textwidth}
        \centering
        \includegraphics[width=\textwidth]{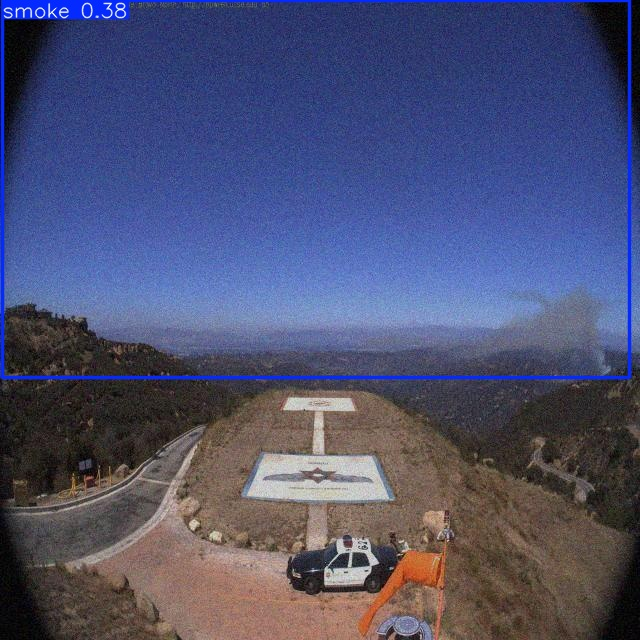} 
        \caption{}
        \label{fig: rtdetr global scene1b}
    \end{subfigure}
    \caption{(a) RT-DETR-l without noise; detection confidence 0.82 and (b) with noise; detection confidence 0.38.}
    \label{fig: rtdetr global scene1}
    \end{minipage}
\end{figure}

Overall, YOLOv8n was more resilient to image-wide perturbation attacks than RT-DETR-l. RT-DETR-l and YOLOv8n had the largest difference in $\mathcal{L}$ at noise level $a=0.210$, where RT-DETR-l's $\mathcal{L}$ was more than 70\% more than that of YOLOv8n ($\mathcal{L}_{\text{RT-DETR-l}}(0.210)=86.53$ and $\mathcal{L}_{\text{YOLOv8n}}(0.210)=50.82$). To illustrate how the global noise affects object detection performance, Figures~\ref{fig: yolo global scene1} and~\ref{fig: rtdetr global scene1} are representative examples that compare original detection results with that at the noise level $0.210$ for identical images.

To the naked eye, the noise is barely visible, only appearing as a dimming in image luminosity. There was minimal difference between the performance of the two models without noise (compare Figure~\ref{fig: yolo global scene1}b with Figure~\ref{fig: rtdetr global scene1}a). However, it is important to note that both models suffered in precision. YOLOv8n tended to be more conservative with its detections, only detecting smoke with a definitive white color, or in other words, at a later period of the incipient stage (see Figure~\ref{fig: yolo global scene1}b). On the other hand, RT-DETR-l did not have this bias but instead began to confuse the sky for smoke with relatively high confidence (see Figure~\ref{fig: rtdetr global scene1}b).

\begin{figure}[hbt]
    \centering
    \begin{minipage}{0.47\textwidth}
    \centering
    \begin{subfigure}[b]{0.48\textwidth}
        \centering
        \includegraphics[width=\textwidth]{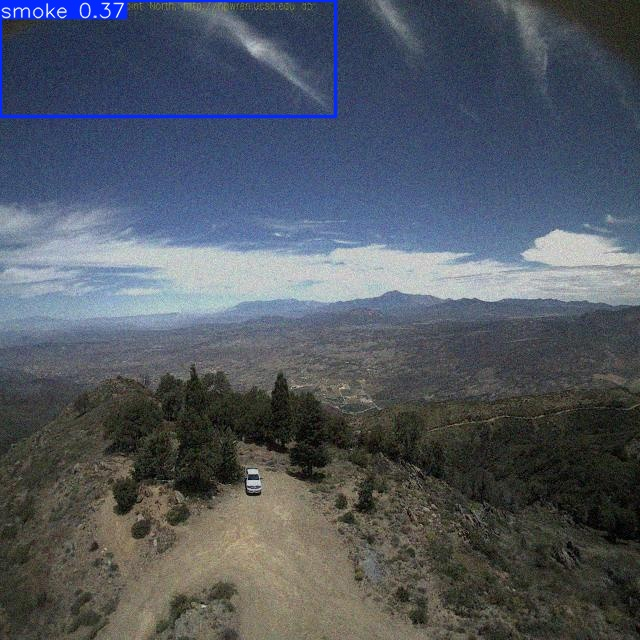}
        \caption{False positive}
        \label{fig: noise global comparison1a}
    \end{subfigure}
    \begin{subfigure}[b]{0.48\textwidth}
        \centering
        \includegraphics[width=\textwidth]{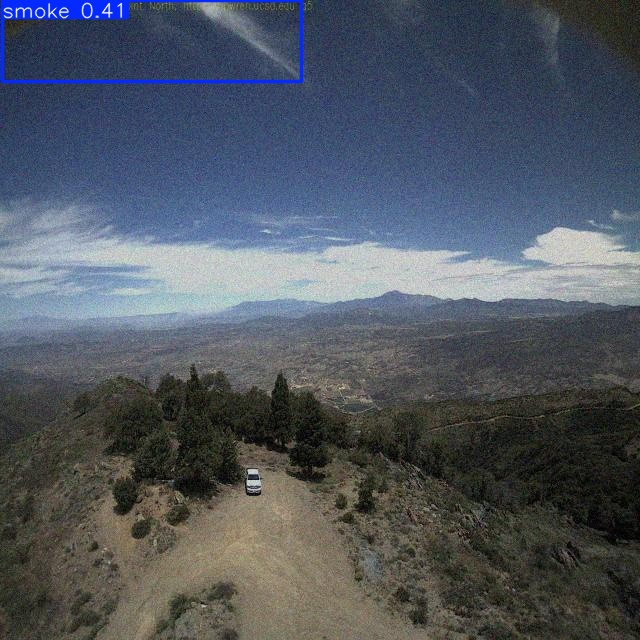} 
        \caption{False positive}
        \label{fig: noise global comparison1b}
    \end{subfigure}
    \caption{(a) YOLOv8n; detection confidence 0.37 and (b) RT-DETR-l; detection confidence 0.41.}
    \label{fig: noise global comparison1}
    \end{minipage}
    \hspace{4mm}
    \begin{minipage}{0.47\textwidth}
    \centering
    \begin{subfigure}[b]{0.48\textwidth}
        \centering
        \includegraphics[width=\textwidth]{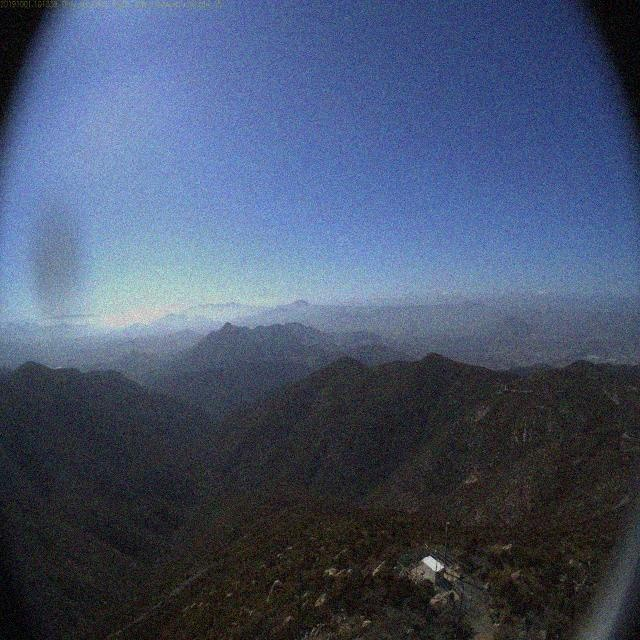} 
        \caption{False negative}
        \label{fig: noise global comparison2a}
    \end{subfigure}
    \begin{subfigure}[b]{0.48\textwidth}
        \centering
        \includegraphics[width=\textwidth]{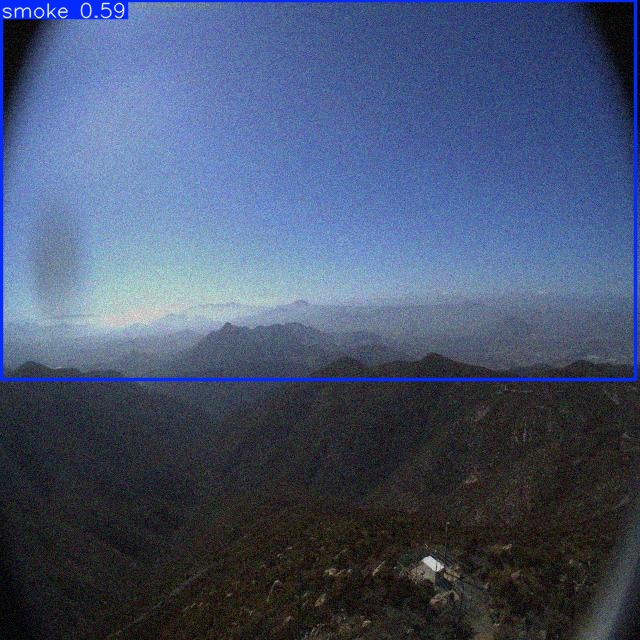} 
        \caption{False positive}
        \label{fig: noise global comparison2b}
    \end{subfigure}
    \caption{(a) YOLOv8n; no smoke detections made and (b) RT-DETR-l; detection confidence 0.59.}
    \label{fig: noise global comparison2}
    \end{minipage}
\end{figure}

Further observation shows that both models may confuse clouds with smoke under noise stress (compare Figure~\ref{fig: noise global comparison1}a with Figure~\ref{fig: noise global comparison1}b), but RT-DETR-l makes false positives even when no clouds exist (see Figure~\ref{fig: noise global comparison2}b). Again, YOLOv8n became conservative with its detections, losing detection confidence and sometimes not detecting at all (see Figure~\ref{fig: noise global comparison2}a). 
Figures~\ref{fig: noise global comparison1} and~\ref{fig: noise global comparison2} are two representative examples of false detections made by both models for identical images, again at noise level $a=0.210$.

This raises model robustness concerns when encountering real image-wide noise. A notable possibility is quality of service degradation events, where network latency or data transfer loss causes footage to become heavily distorted, which may be common given that the cameras are located in remote areas. Other possibilities may include various camera debris (for example, a rain mark near the left in Figure~\ref{fig: noise global comparison2}).

\subsection{Results of Local Deception Test}
In this section, we summarize the results of the local deception test, which included an analysis of the classification flip probabilities and the localization deception rate. In addition, we conduct an auxiliary test that contextualizes the role of image annotations in our findings. 

\subsubsection{Result of Classification Flip Probabilities}
\label{subsubsec: classification flip prob}

To see the robustness of the classification outcome under local noise, we calculate the expected value of both $\alpha_i$ and $\beta_i$ over the entire test set for both YOLOv8n and RT-DETR-l~below. 

\begin{align}
    \mathbb{E}[\alpha_i] = \frac{1}{k}\sum_{i=1}^k \alpha_i,
    \quad \quad
    \mathbb{E}[\beta_i] = \frac{1}{k}\sum_{i=1}^k \beta_i,
\end{align}
where $k$ is the total number of images in the test dataset, which was 1,661 in the HPWREN data (See Table \ref{tab: data split}).

The results are summarized in Table \ref{tab: alpha beta results}. As discussed, $\alpha_i$ shows the robustness of the prediction that image $i$ is smoke-positive, and $\beta_i$ shows the robustness of the prediction that image $i$ is smoke-negative. The results indicate a trend in wildfire smoke detection. Since smoke's features are subtle, the smoke-positive prediction is easily flipped by injecting a cloud-like noise into even one of the 625 grids. This strongly suggests that there is substantial room for improvement in the model, possibly through \textit{data augmentation} with additional images containing smoke-like objects, as discussed in Section \ref{sec: Discussion}. On the other hand, both models show strong robustness for smoke-free predictions. In particular, no smoke-to-null flips were observed for RT-DETR-l, suggesting a greater potential of the transformer-based model for general object detection tasks. 

\begin{table}[htb]
    \centering
    \caption{The Classification Flip Probabilities alongside the true positive/false negative counts without noise.}
    \label{tab: alpha beta results}
    \begin{tabular}{c | c c | c c}
    \hline \hline 
         & $\mathbb{E}[\alpha_i]$ & $\mathbb{E}[\beta_i]$ & TP & FN\\
    \hline 
     \textbf{YOLOv8n}    & $8.67 \times 10^{-6}$ & $0.495$ & $1496$& $165$\\
     \textbf{RT-DETR-l}  & $0.00$ & $0.337$ & $1102$& $559$\\
     \hline
    \end{tabular}
\end{table}

Here, RT-DETR-l's self-attention mechanism may work in its favor for local perturbation \textit{classification}. The architecture may help defend against classification changes under local perturbations, as demonstrated in Table \ref{tab: alpha beta results}. This may be attributable to the self-attention mechanism in transformers, which leverages global relationships across the image. Unlike Section~\ref{subsec: results of global sanity check}, where the entire image was perturbed, local perturbations may not substantially affect the global relationships. 

\subsubsection{Results of Localized Deception Rate}
\label{subsubsec: localized deception rate}

Table~\ref{tab: noise patch sensitivity test} shows the frequency of observed $\gamma_i$ values. We also calculated the expected deception rate.
\allowdisplaybreaks
\begin{align}
\centering
\mathbb{E}_\text{YOLOv8n}\left[\gamma_i\right]
&=
\frac{(0\times1629) + (0.0016\times32)}{1661}=
3.08\times10^{-5} \label{eq: expect for yolo}
\\
\mathbb{E}_\text{RT-DETR-l}\left[\gamma_i\right]
&=
\frac{(0\times1570) + (0.0016\times85) + (0.0032\times5) + (0.0048\times1)}{1661}=
9.44\times10^{-5} \label{eq: expect for rtdetr}
\end{align}

Contrary to Section~\ref{subsubsec: classification flip prob}, RT-DETR-l's self-attention may work to its deficit in local perturbation \textit{localization}. Section~\ref{subsec: object detection framework} demonstrated that RT-DETR's transformer architecture may cause challenges in small-object detection. In particular, object queries representing smoke's subtle features may be de-prioritized, leading to worse precision for small objects. This causes bounding box confusion with similar small smoke-like objects like the cloud PNG patch. This is consistent with Table~\ref{tab: noise patch sensitivity test}, as RT-DETR-l frequently recorded higher $\gamma_i$ values, and its localization deception rate was more than 3 times that of YOLOv8n. However, more testing is required to determine the practical significance.

\begin{table}[bth]
    \centering
    \caption{The frequency of detections for each observed value of $\gamma_i$. Other $\gamma_i$ values were not observed.}
    \label{tab: noise patch sensitivity test}
    \begin{tabular}{c | c c c c}
    \hline 
    \hline
    \textbf{$\gamma_i$} & 0.0000 & 0.0016 & 0.0032 & 0.0048 \\
    \hline
    \textbf{YOLOv8n} & 1629 & 32 & 0 & 0 \\
    \hline
    \textbf{RT-DETR-l} & 1570 & 85 & 5 & 1 \\
    \hline
    \end{tabular}
\end{table}

\begin{figure}[bht]
    \centering
    \begin{minipage}{0.47\textwidth}
    \captionsetup{width=0.9\textwidth} 
    \centering
    \begin{subfigure}[b]{0.48\textwidth}
        \captionsetup{width=0.6\textwidth} 
        \centering
        \includegraphics[width=\textwidth]{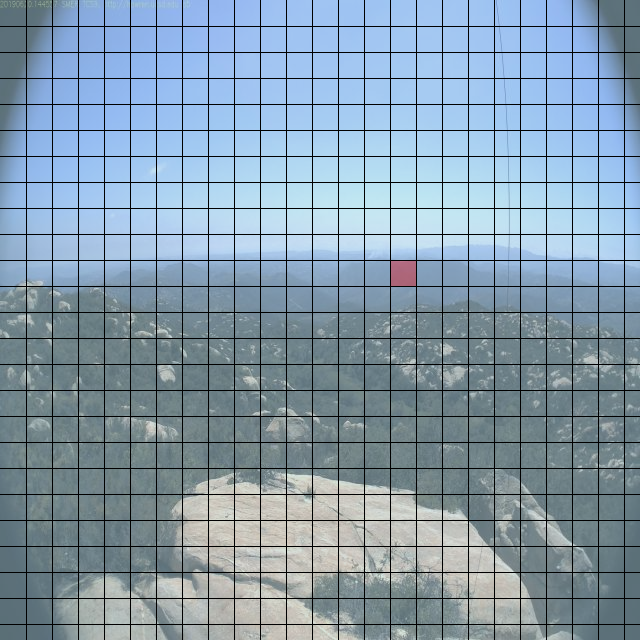}
        \caption{YOLOv8n: $\gamma_{i}=0.0016$}
        \label{fig: noise patch ex1a}
    \end{subfigure}
    \begin{subfigure}[b]{0.48\textwidth}
        \captionsetup{width=0.6\textwidth} 
        \centering
        \includegraphics[width=\textwidth]{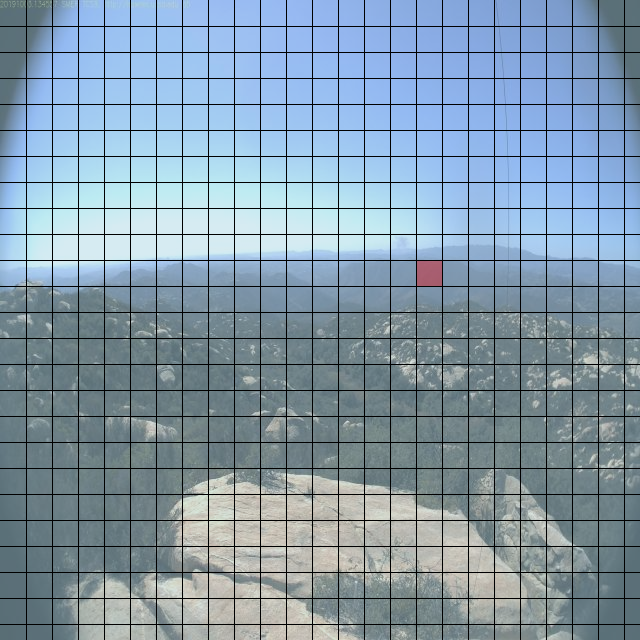} 
        \caption{RT-DETR-l: $\gamma_{i}=0.0016$}
        \label{fig: noise patch ex1b}
    \end{subfigure}
    \caption{SMER TCS9 site (10/3/2019).} 
    \label{fig: noise patch ex1}
    \end{minipage}
    \hspace{4mm}
    \begin{minipage}{0.48\textwidth}
    \captionsetup{width=0.9\textwidth} 
    \centering
    \begin{subfigure}[b]{0.48\textwidth}
        \captionsetup{width=0.6\textwidth} 
        \centering
        \includegraphics[width=\textwidth]{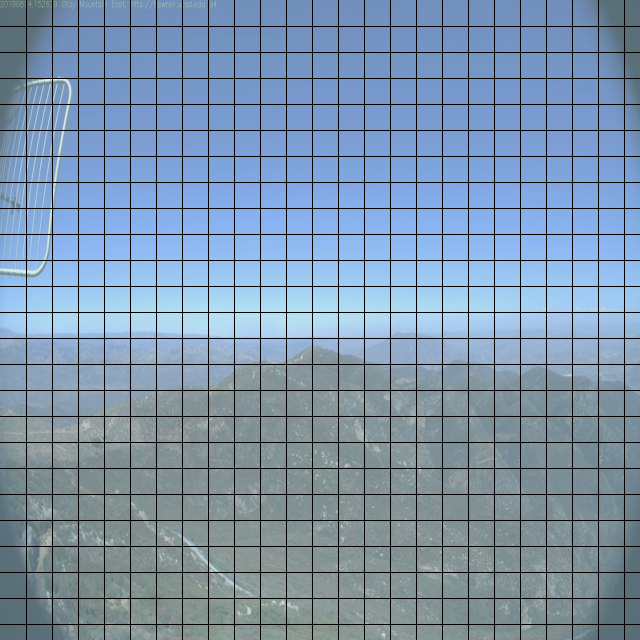} 
        \caption{YOLOv8n: $\gamma_{i}=0.0000$}
        \label{fig: noise patch ex2a}
    \end{subfigure}
    \begin{subfigure}[b]{0.48\textwidth}
        \captionsetup{width=0.6\textwidth} 
        \centering
        \includegraphics[width=\textwidth]{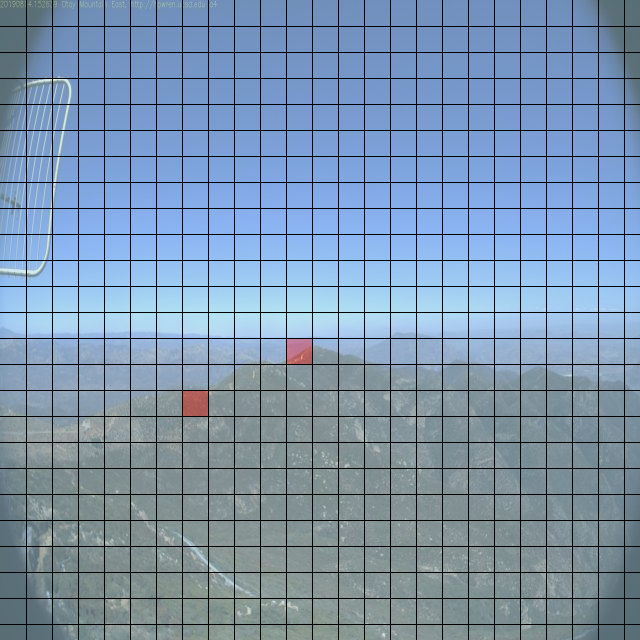} 
        \caption{RT-DETR-l: $\gamma_{i}=0.0032$}
        \label{fig: noise patch ex2b}
    \end{subfigure}
    \caption{Otay Mountain site (8/14/2019).} 
    \label{fig: noise patch ex2}
    \end{minipage}
\end{figure}

\begin{figure}[bht]
    \centering
    \begin{minipage}{0.48\textwidth}
    \captionsetup{width=0.9\textwidth} 
    \centering
    \begin{subfigure}[b]{0.48\textwidth}
        \captionsetup{width=0.6\textwidth} 
        \centering
        \includegraphics[width=\textwidth]{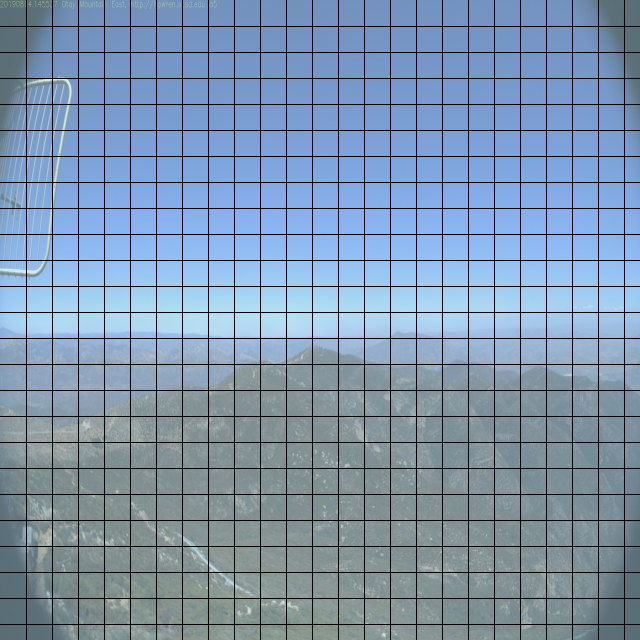}
        \caption{YOLOv8n: $\gamma_{i}=0.0000$}
        \label{fig: noise patch ex3a}
    \end{subfigure}
    \begin{subfigure}[b]{0.48\textwidth}
        \captionsetup{width=0.6\textwidth} 
        \centering
        \includegraphics[width=\textwidth]{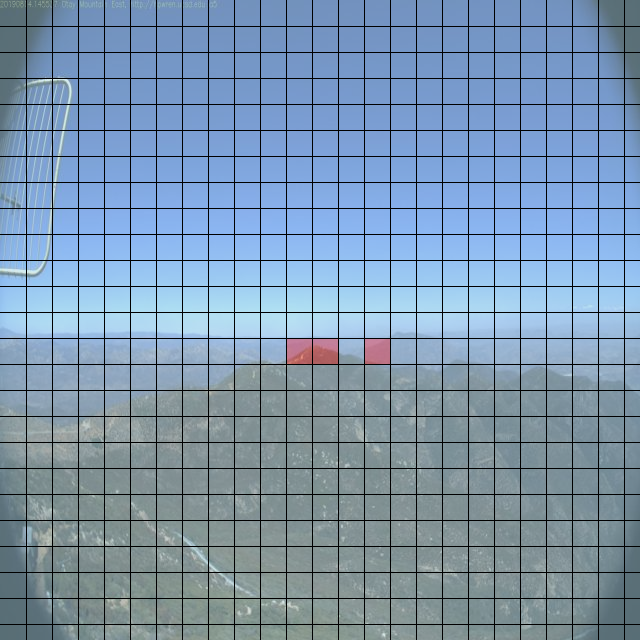} 
        \caption{RT-DETR-l: $\gamma_{i}=0.0048$}
        \label{fig: noise patch ex3b}
    \end{subfigure}
    \caption{Otay Mountain site (8/14/2019).} 
    \label{fig: noise patch ex3}
    \end{minipage}
\end{figure}

For further analysis, we extracted RT-DETR-l detections where $\gamma_{i}>0.0000$, and compared them to their YOLOv8n counterpart. We highlighted where the noise patch deceived the model. There were five images where both models detected with a $\gamma_{i}>0.0000$. Interestingly, four out of five of these detections were taken from the SMER TCS9 surveillance site. Figure \ref{fig: noise patch ex1}a,b show two such cases. There were several cases where RT-DETR-l recorded multiple deceptions whereas YOLOv8n remained completely unaffected. In particular, there were cases in which RT-DETR-l detected with a $\gamma_{i}$ value of 0.0032 and 0.0048 (see Figures~\ref{fig: noise patch ex2}b and \ref{fig: noise patch ex3}b, respectively), but YOLOv8n remained unchanged and detected with $\gamma_{i}$ value of 0.0000 (see Figures \ref{fig: noise patch ex2}a and \ref{fig: noise patch ex3}a). Below are identical images detected by the two models, where each grid shows where the cloud-noise was injected. Red areas indicate noise-affected regions.

Factoring out rare cases where the cloud noise completely overlapped the smoke (this could only happen in one or two grids, putting the probability at around 0.16--0.32\%), both models were deceived when the cloud PNG patch was near the smoke. This is alarming, since the cameras are positioned at vantage points, and clouds are more clearly visible at higher altitudes. Thus, the occlusion of smoke by clouds would be a relatively common occurrence. Furthermore, it can be explained that the models confuse smoke with cloud since both are similar in features, and the parts of the cloud PNG patch merge with the smoke at times. As seen with YOLOv8n, the models tend to reliably detect when the smoke has developed a clear white coloration. Thus, the models may have interpreted the cloud as the origin point of the smoke, even when the direction of the smoke plumes suggested otherwise. 

\begin{figure}[hbt]
    \centering
    \begin{subfigure}[b]{0.3\textwidth}
        \centering
        \includegraphics[width=\textwidth]{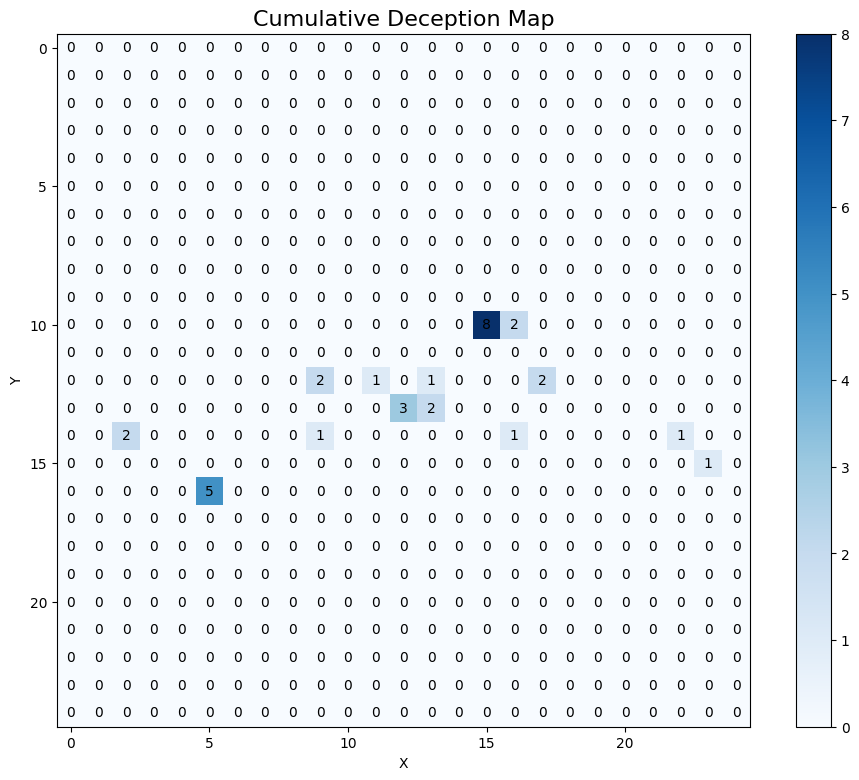}
        \caption{81.25\% of deceptions within the middle-horizontal (20\% of image).}
        \label{fig: cumul decep map a}
    \end{subfigure}
    \hspace{2cm}
    \begin{subfigure}[b]{0.3\textwidth}
        \centering
        \includegraphics[width=\textwidth]{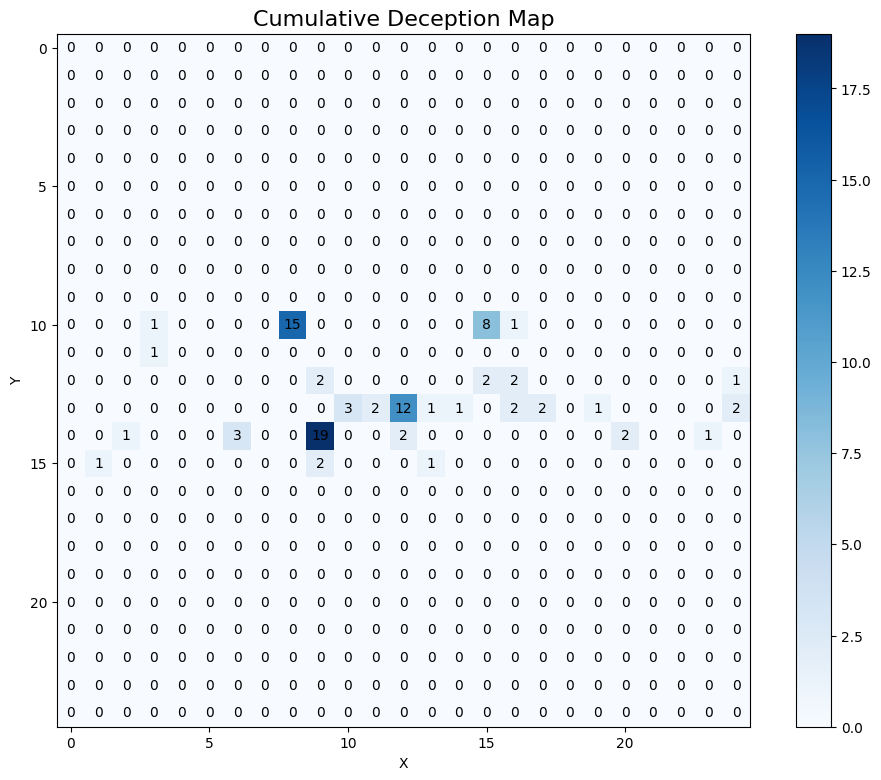}
        \caption{88\% of deceptions within the middle-horizontal (20\% of image).}
        \label{fig: cumul decep map b}
    \end{subfigure}
    \caption{Cumulative Deception Maps for YOLOv8n (a) and RT-DETR-l (b).}
    \label{fig: cumul decep map}
\end{figure}

\begin{figure}[hbt!]
    \centering
    \colorbox{gray!5}{ 
        \includegraphics[width=0.3\linewidth]{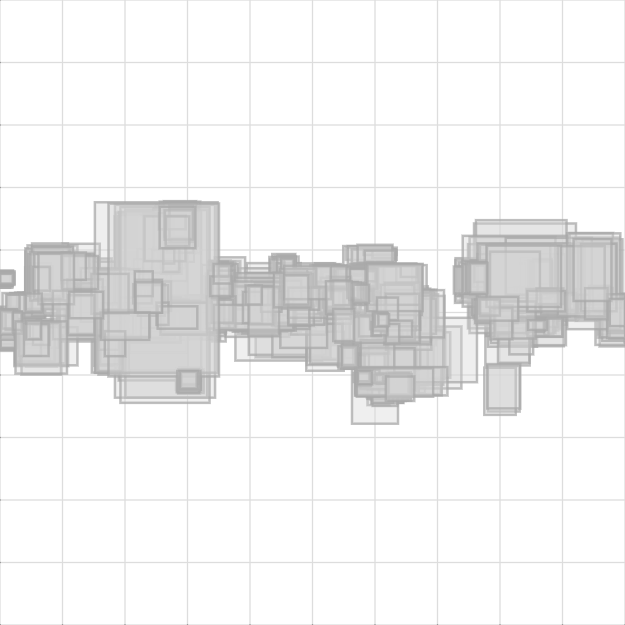}
    }
    \caption{Distribution of annotations in the test dataset.}
    \label{fig: annotation heatmap}
\end{figure}

To test if there was a spatial relationship between the deceptions, we enumerated the cumulative number of noise-affected detections for each slot that the patch noise was injected for both models. Both models exhibited a slight bias to be deceived at the center regions (see Figure~\ref{fig: cumul decep map}a,b). A potential explanation is that smoke is most frequently depicted in the middle horizontal of the image. It becomes easier for human observers to spot and annotate as smoke rises above the horizon, which is frequently in this region. {This creates an annotation bias in the test set (see Figure}~\ref{fig: annotation heatmap}{), which may have caused the models to over-focus on this region.} Thus, cloud PNG patches may appear as part of the smoke most in this zone.

\section{Discussion}
\label{sec: Discussion}

In a comprehensive comparative study, {we analyze} the adversarial robustness of CNN- and transformer-based real-time smoke detection models {in context to other studies}. The main findings are summarized below:

\begin{itemize}
    \item The global sanity check revealed that the transformer-based RT-DETR-l model was substantially more vulnerable to global noise injection compared to the CNN-based YOLOv8, even at noise levels barely visible to the human eye. {This is a notable contrast from other studies}~\citep{mahmood2021robustness, shao2021adversarial}~which suggest that transformers trained on Big Data
    (i.e., CIFAR-10, MNIST, etc.)  are more robust than CNNs against global noise. The current results may be an artifact of the severe data shortage in wildfire detection. Thus, as more data augmentation techniques (such as those from WARP) create more diverse adversarial examples, transformers will most likely overtake CNNs in terms of global noise robustness.
    \item An analysis of the classification flip probabilities revealed that both CNN- and transformer-based models are sensitive to local noise injection. A single noise--injected grid (out of 625 total) resulted in flip probabilities of $\approx$50\% and $\approx$34\% in the smoke-positive prediction for the CNN and transformer models, respectively. These results underscore the need for further model training using data augmentation techniques. {While transformers performed better during this test in our study,}~\cite{fu2022patch,gu2022vision}~{suggest that transformers are not necessarily stronger at patch perturbations than CNNs. Because this seems to be a continuity even for Big Data models, more research is recommended for this adversarial attack.}
    \item An analysis of the localization deception rate revealed areas within the images that were particularly vulnerable to local noise injection. {Detailed analysis suggests that human annotation bias may cause to over-focus on the middle-horizontal region, offering insights for future data augmentation strategies to enhance model robustness. This is consistent with DNN behavior in other fields}~\citep{dhar2022systematic},~{highlighting the need to consider not only the visual characteristics of the target objects but also its spatiotemporal context for true unbiased training.}
\end{itemize}

Based on our analysis, we propose the following data augmentation strategies {for improving robustness. These solutions should diversify future training data for YOLOv8- and RT-DETR-based wildfire detection models. Moreover, because this framework is model-agnostic, these solutions may still be applied even if new DNN architectures are introduced.}

\label{desc: future work}
\begin{itemize}
    \item Gaussian-distributed Noise 

    \begin{itemize}
        \item {
        Gaussian noise with $0.1 \lesssim a \lesssim 0.4$ should be introduced into training data, which should reduce precision degradation when encountering global noise, especially among speed-optimized transformers. It will also improve data variety and quantity. }
    \end{itemize}

    \item Cloud PNG-Patch 

    \begin{itemize}
        \item{
        Clouds and smoke overlap most in the middle-horizontal strip of the images. Given that most false positives occur in this zone, cloud PNG patches should be placed in this zone to help models distinguish the two.}
        \item Furthermore, cloud PNG patches should also be placed into the upper areas (areas depicting the sky) to add spatial variety to local noise patches.
    \end{itemize}

    \item Collages/Mosaic
    
    \begin{itemize}
        \item An effective solution to combat false positives extensively used by \cite{yazdi2022nemo} is to use collages. This is because collages {allow models to easily compare} between smoke-positive and smoke-negative images. Since CNNs suffered from local perturbation \textit{classification}, collages of images between classes should be implemented into training data.
        \item Furthermore, {certain collage techniques such as the} YOLOv4-style mosaic \citep{bochkovskiy2020yolov4} has the added benefit of introducing object size variety in the data. This is particularly useful for small object detection. 
        \item However, collages inevitably make the already-small smoke object even smaller. Especially for speed-optimized transformers, which have a known weakness to small objects, crucial smoke features must be extracted first. The below augmentation strategy seeks to offer a potential solution. 
    \end{itemize}

    \item $\mathbf{2\times2}$ Crops
    \begin{itemize}
        \item Data collected by \cite{govil2020preliminary} from \citep{HPWREN} exclusively depict smoke that appears at or near the horizon (i.e., the middle-horizontal of the image). {Cropping these images} into equal quadrants{ creates four standalone images, which shift the position of objects. This adds spatial diversity to smoke annotations.}
        \item Furthermore, since the crops will result in a smaller-sized image, when resized to the target image size ($640\times640$ pixels), the {image, and by extension, smoke,} will appear larger. This may help models better extract the subtle features of smoke, offering a solution to the problem discussed in the previous augmentation strategy. 
        \item Finally, crops that do not include smoke introduce negative samples. Generally speaking, negative samples will help reduce false positives.
    \end{itemize}
\end{itemize}


\section{Conclusions}

In this study, we introduced WARP
 (\textbf{W}ildfire \textbf{A}dversarial \textbf{R}obustness \textbf{P}rocedure), 
 the first-ever model-agnostic framework for evaluating the adversarial robustness of wildfire smoke detection models, designed to address limitations arising from insufficient variety in smoke images.

WARP supports both global and local adversarial attack methods. While the global attack method employs image-contextualized random noise overlays, the local attack method is tailored to address two key aspects of smoke detection: (1) classification between smoke-positive and smoke-negative instances and (2) smoke object bounding box localization.

{Leveraging WARP's model agnostic capabilities, we trained and compared CNN- and transformer-based wildfire detection models. We found in the global attack method that transformers suffer from higher precision degradation under global noise. In the local attack method, we found that while CNNs' classification results are less robust than those of transformers, CNNs performed better in terms of localization ability. Future studies regarding this perturbation method are recommended to obtain a conclusive result. Finally, an auxiliary test suggests that annotation bias may be partially responsible for deceiving both models with local noise patches.} Based on these findings, we proposed wildfire-specific data augmentation approaches. We leave a detailed analysis of the proposed data augmentation approaches for future studies.

{WARP is a large step towards making DNN-based wildfire detection models more practical. Full-scale implementation of these systems will inevitably be costly. With the already high firefighting costs in the US, vulnerabilities to even simple adversarial attacks add great uncertainty in the reliability of these models. By identifying these biases and creating simple countermeasures in the form of data augmentation, WARP ensures both model accountability and dependability. We hope that this work will advance wildfire detection models enough to fully implement a completely automatic prediction system against wildfires, which can save the valuable lives and infrastructure that are lost to it.}


\section*{Acknowledgements}
We would like to thank Dr. Amirhesam Yazdi from the University of Nevada Reno for their thoughtful discussions, alongside Mr. Patrick Watters's generous help for debugging. We also acknowledges Mr. Austin Parkerson for his continued support in providing the GPU computing service which was vital for the experimentation.


\section*{Data and Code Availability Statement}
The code used in this study has been open-sourced and can be found in the official \href{https://github.com/ri2658/WARP.git}{WARP GitHub repository}. Moreover, the data used in this study are openly available from \cite{yazdi2022nemo} (Apache 2.0 License) and \cite{govil2020preliminary} (CC BY-NC-SA 4.0 Licence).

\appendix
\section{Appendix}
\subsection{Definition of mAP}
\label{subsec: map}

Mean Average Precision (mAP) 
 is a common metric for precision in machine learning. It is obtained by calculating numerous other sub-metrics, which are shown here. 

Precision 
 and Recall 
 are well established performance metrics in machine learning. Average Precision 
 can be calculated by taking the area beneath the Precision--Recall curve (i.e., $p(r)$), typically using the Trapezoid Rule. 
\begin{align}
Precision &=\frac{TP}{TP+FP},\\
Recall &=\frac{TP}{TP+FN},\\
AP&=\int_{0}^{1}p(r)dr,
\label{eq: ap}
\end{align}
where TP is the number of true positives, FP is the number of false positives, and FN is the number of false negatives. A detection is considered a true positive when the detection bounding box $A$ overlaps the ground truth bounding box $B$ by a certain overlap threshold. 

The degree of overlap is quantified by the Intersection over Union (IoU) 
, which is obtained by 

\begin{equation}
IoU(A,B)=\frac{\left| A\cap B \right|}{\left| A\cup B \right|}.
\label{eq: IoU}
\end{equation}

The \textit{mean Average Precision (mAP)} can be obtained by taking the average of AP across all classes. 
\begin{equation}
mAP=\frac{\sum_{l=1}^{k}AP(l)}{k},
\label{eq: map}
\end{equation}
where the class index $l$ runs over all the class labels $\{1,\ldots, k\}$ with $k$ being the total number of classes. 

There are certain variations of mAP based on what IoU threshold is used. The most common is \textit{mAP50:95} or simply \textit{mAP}, which is the average of mAP scores from thresholds 50 to 95 at increments of 0.05. There is also \textit{mAP50}, where mAP is calculated with a fixed overlap threshold of $IoU=50$.

\bibliographystyle{apalike}  
\bibliography{bib}

\end{document}

%% file: figures/example_figure.tex
\definecolor{neon}{rgb}{0.3, 1.0, 0.3}
    \begin{tikzpicture}
        \node (img1) at (0,0) {\includegraphics[width=0.3\textwidth]{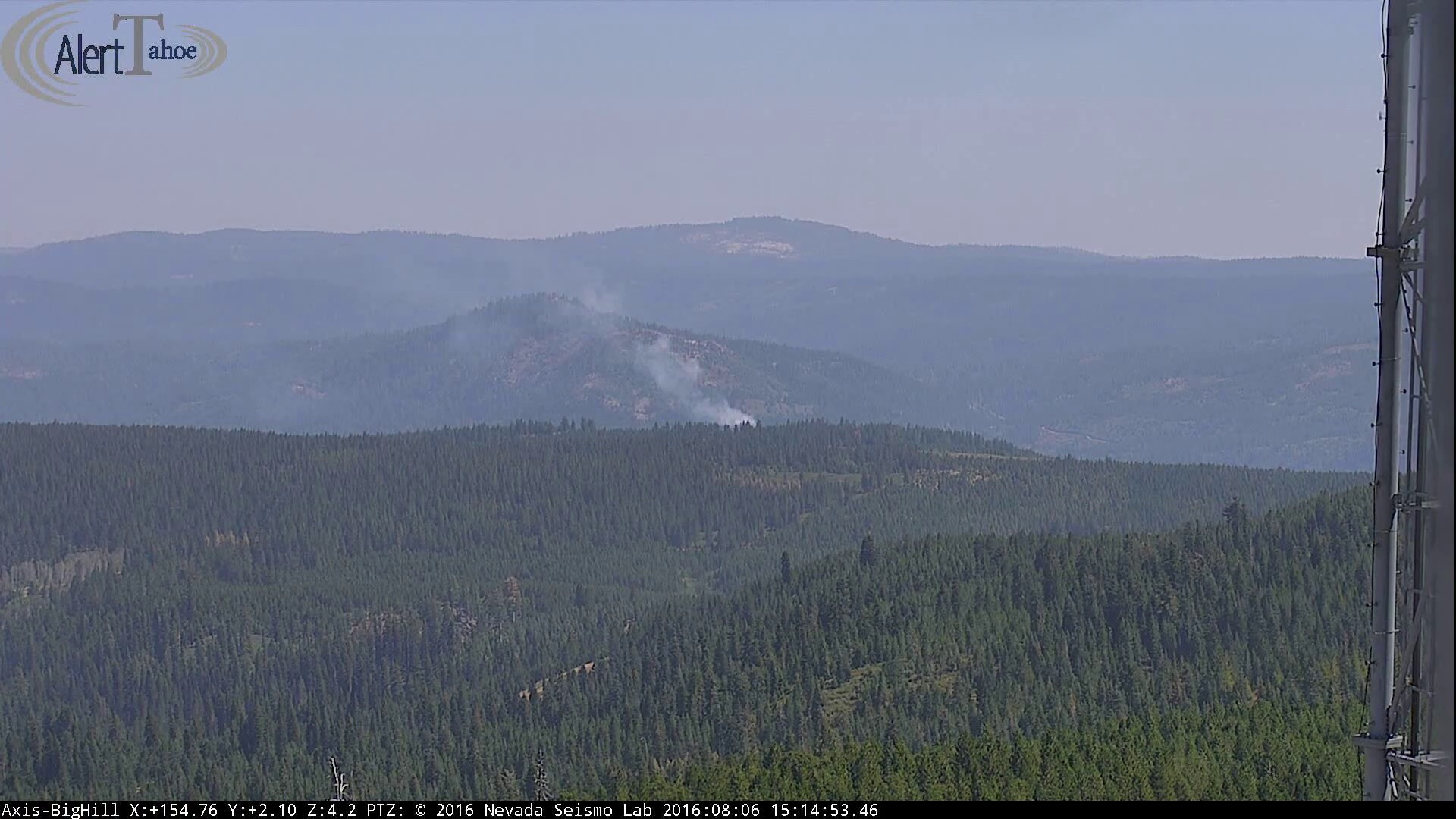}};
        
        \node (img2) at (8,0) {\includegraphics[width=0.3\textwidth]{figures/example_image.png}};
        
        \draw[->, very thick] (3,0) -- (5,0) 
            node[midway, below, yshift=-0.5cm] {\shortstack{DNN-Object \\ Detection Model}};

        \draw[neon, line width=1pt] (7.1, -0.09) rectangle ++(1.05,0.6);
    \end{tikzpicture}

%% file: figures/noise_workflow.tex
\definecolor{lightGreen}{RGB}{217,234,211}
\definecolor{brightRed}{RGB}{255,0,0}
\definecolor{brightBlue}{RGB}{0,0,255}


\begin{tikzpicture}
    \centering
    \draw[draw=black, fill=lightGreen] (0,0) rectangle (4,4);
    \draw (0,0) grid (4,4);
    \node (img1) at (0.5,3.5) {\includegraphics[height=0.075\textwidth, width=0.075\textwidth]{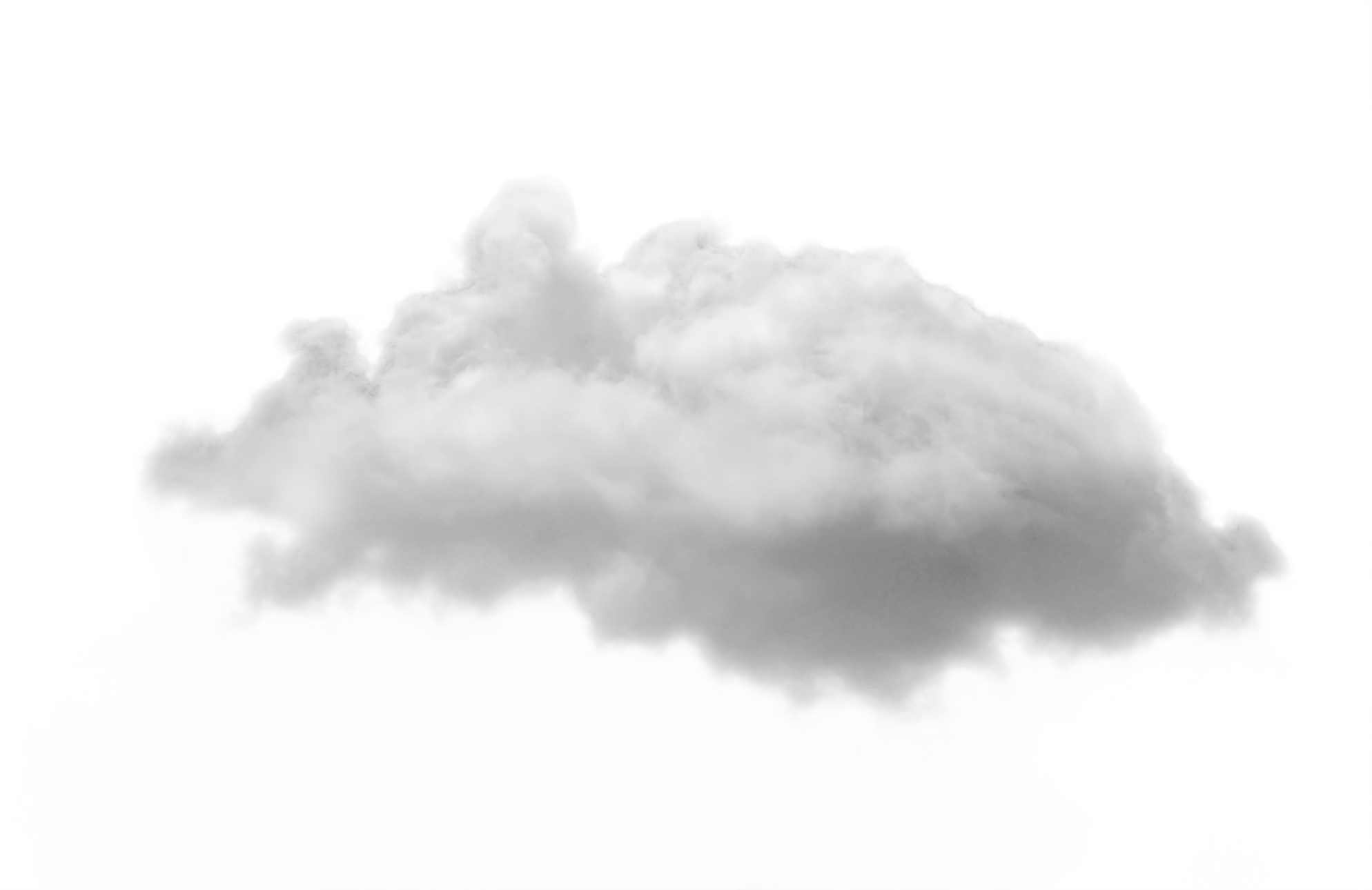}};
    \node (img2) at (2.5,1.5) {\includegraphics[height=0.075\textwidth, width=0.075\textwidth]{figures/cloud.png}};
    \draw (2.5,1.5) -- (4.5,2.5);
    \draw[draw=brightRed] (0.1,3.2) rectangle (0.9,3.8);
    \draw[draw=brightRed] (2.1,1.2) rectangle (2.9,1.8);
    \draw[draw=brightBlue] (2.25,1.05) rectangle (3.05,1.65);
    \draw[->] (0.5,3.2) -- (0.5,2.7);
    \draw[->] (0.9,3.5) -- (1.4,3.5);
    \node at (2, 4.25) {Image Index $i$};
    \node[rotate=90] at (-0.25, 2) {$n$ slots};
    \node at (2, -0.25) {$n$ slots};
    \node at (2, -1) {Attempts: $A_{i}=n\times n=625$};
    \node[align=center] at (7,2.5) {$IoU(\text{BBox}_{\text{patch}},\text{BBox}_{\text{detect}})\geq 0.50$\\
    $\# (\text{Deceived Detections})=D_{i}$
    };
    \node[align=center] at (7,0.5) {\textbf{Deception Rate} \\ 
    $\boldsymbol{\gamma_{i}=\frac{D_{i}}{A_{i}}}$};
\end{tikzpicture}
